\documentclass[letterpaper, 10 pt, conference]{ieeeconf}  %

\IEEEoverridecommandlockouts                              %

\usepackage{graphics} %
\usepackage{times} %
\usepackage{amsmath} %
\usepackage{amssymb}  %
\usepackage{amsfonts}
\usepackage{subfigure}
\usepackage{hyperref}
\usepackage{graphicx}
\usepackage{float}
\usepackage{caption}
\usepackage{multirow}
\usepackage{booktabs}
\usepackage{color}

\title{\LARGE \bf
Differentiable Fluid Physics Parameter Identification Via Stirring
}

\author{Wenqiang Xu$^{*1}$, Dongzhe Zheng$^{*2}$, Yutong Li$^{1}$, Jieji Ren$^{1}$, Cewu Lu$^{1}$%
\thanks{*Equal contribution}%
\thanks{$^{1}${\tt\small \{vinjohn, davidliyutong, jiejiren, lucewu\}@sjtu.edu.cn}. Cewu Lu is the corresponding author, a member of Qing Yuan Research Institute and MoE Key Lab of Artificial Intelligence, AI Institute, Shanghai Jiao Tong University, China.}%
\thanks{$^{2}${\tt\small dz1011@wildcats.unh.edu}. Work is done when Dongzhe is an intern at SJTU.}%
}

\begin{document}

\maketitle
\thispagestyle{empty}
\pagestyle{empty}

\begin{abstract}
Fluid interactions permeate daily human activities, with properties like density and viscosity playing pivotal roles in household tasks. While density estimation is straightforward through Archimedes' principle, viscosity poses a more intricate challenge, especially given the varied behaviors of Newtonian and non-Newtonian fluids. These fluids, which differ in their stress-strain relationships, are delineated by specific constitutive models such as the Carreau, Cross, and Herschel-Bulkley models, each possessing unique viscosity parameters. This study introduces a novel differentiable fitting framework, DiffStir, tailored to identify key physics parameters via the common daily operation of stirring. By employing a robotic arm for stirring and harnessing a differentiable Material Point Method (diffMPM)-based simulator, the framework can determine fluid parameters by matching observations from both the simulator and the real world. Recognizing the distinct preferences of the aforementioned constitutive models for specific fluids, an online strategy was adopted to adaptively select the most fitting model based on real-world data. Additionally, we propose a refining neural network to bridge the sim-to-real gap and mitigate sensor noise-induced inaccuracies. Comprehensive experiments were conducted to validate the efficacy of DiffStir, showcasing its precision in parameter estimation when benchmarked against reported literature values. More experiments and videos can be found in the supplementary materials and on the website: \url{https://sites.google.com/view/diffstir}.
\end{abstract}

\section{INTRODUCTION}
Humans interact with fluid every day, from mixing up coffee powder into water to squeezing toothpaste. Understanding fluid is the key to state estimation in a larger range of fluid-related manipulation tasks. In domestic scenarios, density and viscosity are two important physics properties to identify. For example, in the kitchen, adjusting density usually relates to adjusting the extent of a certain flavor (e.g., sweetness and saltness). Viscosity can indicate the solidity of the pastry after the bakery. Other physics properties such as elasticity also contribute to the fluid dynamics behaviors but do not have a significant influence on household tasks, thus they are not in our major focus. Specifically, common fluid has a bulk modulus to characterize the elasticity which ranges from 0.9GPa to 4.5GPa \cite{EngineeringToolboxBulks}. It means at least $9$MPa should be applied to the fluid to make it compress by 1\% volume, such pressure is unusual in household tasks.

\begin{figure}[t!]
    \centering
    \includegraphics[width=1\linewidth]{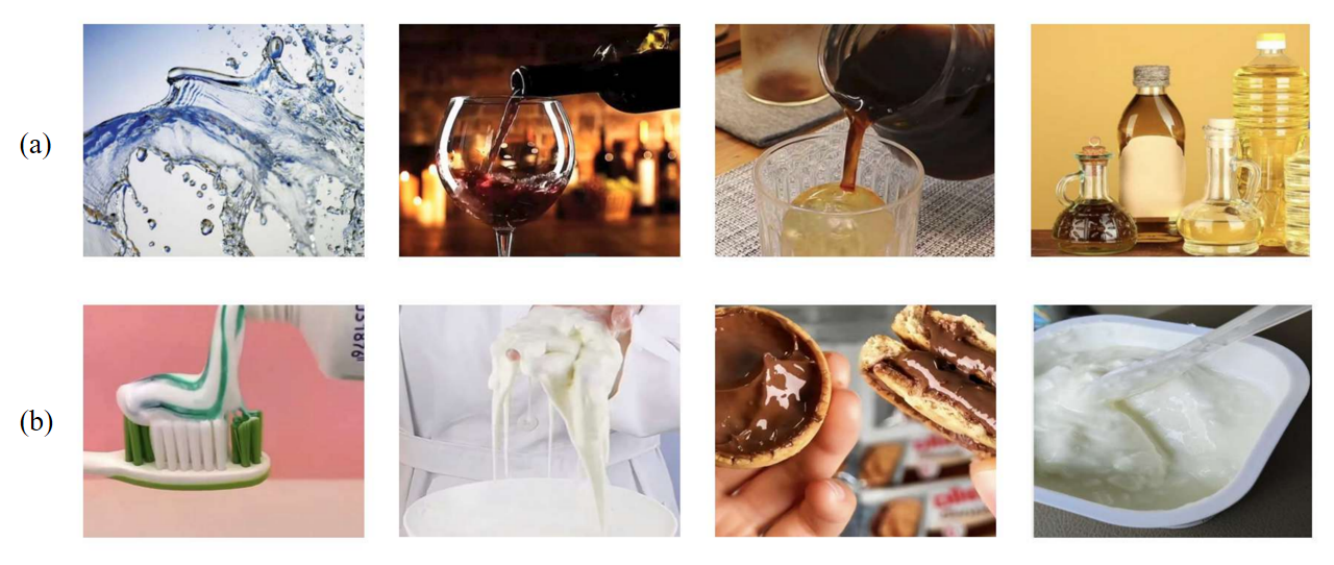}
    \caption{(a) Newtonian fluids: distilled water, red wine, black coffee, and cooking oil. Despite their varied molecular structures, fluids in (a) consistently obey Newtonian principles. (b) Non-Newtonian fluids and their properties: toothpaste (Bingham plasticity), Oobleck (shear-thickening, impact resistance), chocolate sauce (thixotropy), and Bulgarian yogurt (viscoelasticity).}

    \label{fig:liquids_example}
\end{figure}

Though density can be easily estimated with Archimedes' principle, viscosity is more complex. In hydraulic mechanics, the dynamics behaviors of fluids are described by different constitutive models \cite{harris2005fast, anderson1995computational, park2023fluid}, which means they can be characterized by a few parameters given the predefined stress-strain relationship. Based on the stress-strain relationship, fluids can be categorized in \textit{Newtonian} (Fig. \ref{fig:liquids_example}-a) and \textit{non-Newtonian} (Fig. \ref{fig:liquids_example}-b) types. Newtonian fluids are characterized by a linear stress-strain relationship.
While non-Newtonian fluids have varied viscosity based on the rate of deformation caused by the shear stress. These fluids are typically described by constitutive models such as the Carreau \cite{carreau1972rheological}, Cross \cite{cross1965rheology} and Herschel-Bulkley \cite{herschel1926konsistenzmessungen} models. In these models, viscosity is one of the major parameters. 

In this work, we present a novel differentiable fitting framework named \textbf{DiffStir} to identify the key physics parameters through a simple action, stirring, a common operation in the kitchen. It utilizes a robot arm to grasp a rod and conducts the robotic stirring operation on a cup of certain fluid with recorded trajectories and forces. After synchronizing the action and force between the real-world configuration and a differentiable Material Point Method (diffMPM) \cite{hu2019difftaichi}-based simulator, we can estimate the fluid parameters of the pre-set constitutive models by matching the observations from both worlds. Different non-Newtonian constitutive models have their own preferences for fluid. For example, the Carreau model is suitable for shear-thinning fluids that can approach Newtonian behaviors at high shear rates; the Cross model is good for shear-thinning and shear-thickening fluids across a wide shear rate range; and the Herschel-Bulkley model is best for fluids with a yield stress, such as muds and some toothpaste. Based on it, we design an online strategy to adaptively select the most proper constitutive models given the real-world recordings. Moreover, though MPM is a powerful computational tool often employed in fluid dynamics due to its ability to handle large deformations and topological changes in a numerically stable manner, the physical sim-to-real gap is inevitable, not to mention the real-world sensor noise, which causes inaccuracy of the parameter estimation. To address it, we propose a refining neural network to calibrate the parameters between the real world and the simulator. 

To evaluate the proposed DiffStir, we conduct extensive experiments from four aspects to demonstrate the accuracy of the estimated parameters with reference values from the literature \cite{EngineeringToolboxDynamicViscosities, EngineeringToolboxViscosity, EngineeringToolboxKinematicViscosities}: density validation, density \& apparent viscosity estimation, fluid dynamics behavior classification, and fluid mixing analysis.

We summarize our contribution as:
\begin{itemize}
    \item We propose to identify the key physics parameters, density and viscosity, of fluid in household scenarios through a common operation, stirring. The identification process bridges the real world and the differentiable simulator, and can achieve fluid understanding during robotic operation.
    \item We conduct extensive experiments on different Newtonian and non-Newtonian fluids to validate the usability and accuracy of our system. We hope our system can serve as a foundational tool for liquid-related manipulation tasks.
\end{itemize}

\section{Related Works}

\subsection{Fluid Constitutive Model}
Fluid dynamics research has provided insights into complex phenomena, such as viscoelasticity, viscoplasticity, and elastoplasticity, advancing our understanding of fluid properties and applications.

For viscoelastic models, Maxwell \cite{maxwell_ref} and Oldroyd-B \cite{oldroyd_b_ref} are fundamental in understanding biological and polymeric fluids, relevant to biomedicine and polymer processing.
For elastoplastic models, Prandtl-Reuss \cite{prandtl_ref} and Desai \cite{desai_ref} capture elastoplastic behaviors seen in foams and slurries, proving vital in subsurface and construction studies.
For viscoplastic models, Casson \cite{casson_ref}, Sisko \cite{sisko_ref}, and Bingham \cite{bingham_ref} elucidate viscoplastic behaviors, emphasizing the yield stress in substances such as blood, toothpaste, and mayonnaise.

In our work, we integrate the Carreau, Cross, Herschel-Bulkley \cite{chen2021constitutive}, and Kelvin-Voigt models \cite{dowling1999mechanical} to represent diverse fluid behaviors. Together, they capture shear behaviors, yield stress, and viscoelastic properties, offering a versatile framework suitable for a wide range of applications.

\subsection{Fluid Identification Method}
Fluid parameter identification, encompassing viscosity, and density, utilizes multiple methodologies, such as measurement instruments-based, and computation-based. 

For those methods with measurement instruments. The falling ball viscometer assesses fluid viscosity but is influenced by variables like temperature \cite{brizard2005design, labuza1985effect}. Oscillatory shear tests provide insights into complex fluids but might not always capture certain viscoelastic fluids accurately \cite{gunasekaran2000dynamic}. Rotational viscometry, although popular, occasionally encounters issues with non-Newtonian fluids \cite{powell1998rotational}. Our approach integrates differentiable physics simulation and real-world robotic feedback to augment these conventional techniques.

For those methods with computational algorithms. An important line of work is built upon Computational Fluid Dynamics (CFD) \cite{soares2008analysis,carvalho2023parameter,urbanowicz2022inverse}, they can identify the parameters along with the solution of CFD computation. Yet, its computational demands can hinder interactive robotic applications. Our method takes the essence from CFD, following conservation laws and the Navier-Stokes equations, maintaining the common constitutive relations in CFD, and simplifies the computation by employing a particle-based representation in MLS-MPM for the natural handling of large deformations.
Another line is data-driven-based approaches, which leverage machine learning, particularly neural networks, for fluid parameter determination \cite{fukami2021model,morimoto2021convolutional,reddy2019reduced,belbute2020combining}. Despite their predictive power, some of them lack robust physical validation, potentially inducing errors. Contrarily, our technique emphasizes a robust physical framework, ensuring accuracy and stability.

\subsection{Fluid Simulation for Robot}
Numerous robotic simulation platforms exist, but simulating non-Newtonian fluids remains challenging. While PyBullet \cite{coumans2016pybullet} and MuJoCo \cite{todorov2012mujoco} are prominent in robotic dynamics, they lack fluid simulation capabilities. Subsequent platforms like FleX \cite{macklin2014FleX}, SoftGym \cite{lin2021softgym}, SAPIEN \cite{xiang2020sapien}, and TDW \cite{gan2020TDW} largely focus on Newtonian fluid dynamics. FleX, although advanced, serving as the foundation in the recent Isaac Sim \cite{isaacgym}, is limited by its closed source and scalability. SoftGym emphasizes deformable objects, while SAPIEN, leveraging the PhysX engine \cite{harris2009cuda}, underemphasizes fluid mechanics. TDW offers photorealistic visuals but lacks physics interaction. A common limitation across these platforms is their restriction to Newtonian fluids, neglecting non-Newtonian fluid intricacies prevalent in real robotic scenarios.

Recent differentiable platforms like SPNets \cite{schenck2018spnets} and PHIFLOW \cite{holl2020PhiFlow} have been proposed. SPNets, pioneering in robot-fluid interactions, is limited in single-phase non-Newtonian fluids. PHIFLOW, although excels in fluid simulation, faces challenges when integrating solid interactions.

In contrast, our method, compared to FluidLab, employs the Kelvin-Voigt viscoelastic model to capture nuanced non-Newtonian fluid behaviors, covering effects from shear thinning to thickening. FluidLab, on the other hand, omits essential fluid parameters like viscosity, as well as other complex rheology phenomena such as shear thinning and shear thickening.

\section{Problem Statement}
Fluid dynamics exhibit intricate behaviors with many unresolved issues \cite{davidson2015turbulence, majda2002vorticity}. Different constitutive models are proposed to characterize the dynamics behavior of fluids \cite{chen2021constitutive}. In this work, we exploit the Kelvin-Voigt model \cite{dowling1999mechanical}. 

\begin{figure}[htbp]
    \vspace{0.2cm}
    \centering
    \includegraphics[width=1\linewidth]{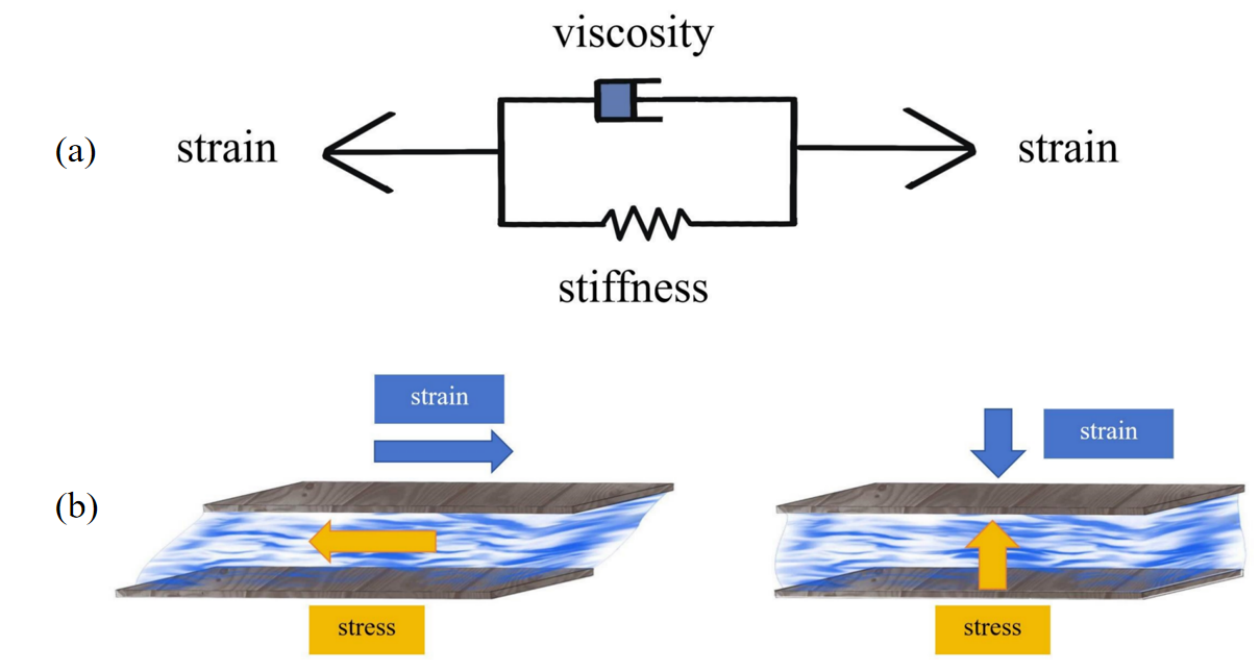}
    \caption{(a) Kelvin-Voigt model considers the viscosity and elasticity of a fluid. The dashpot stands for the fluid's resistance to flow. The spring stands for the fluid's ability to recover its original shape after deformation, as well as its incompressibility property. (b) Two sub-figures representing fluid dynamics: one with relative sliding between plates, focusing on dynamic viscosity and shear resistance, and the other with a downward force on a plate emphasizing bulk modulus and compressibility. %
    }
    \label{fig:constitutive}
    \vspace{-0.2cm}
\end{figure}

\textit{1) Kelvin-Voigt Fluid Model:}
The Kelvin-Voigt model, merging a spring (representing elasticity) and a dashpot (representing viscosity) in parallel (Fig. \ref{fig:constitutive}(a)), provides the rheological characteristics of diverse viscoelastic materials, such as polymeric solutions, foams, gels, and suspensions \cite{meyers2008mechanical}. The model derives the total stress, $\sigma$, from the collective contributions of both the viscous and the elastic elements:
\begin{equation}
    \sigma = E \varepsilon + \eta \dot{\varepsilon},
\end{equation}
where $\varepsilon$ is the strain applied to the fluid, $E$ is the elastic modulus to characterize the stress-strain relationship of the spring, $\dot{\varepsilon}$ is the strain rate and $\eta$ is the viscosity parameter.

\textit{2) Integrating Volume-Related Energy Density:}
The Kelvin-Voigt model utilizes a bulk modulus $\kappa$ to quantify the elastic response to volume change. Since the fluid undergoes non-linear elastic deformation, we employ the Neo-Hookean model, which %
details the non-linear elastic reaction to volume changes through the energy density function \(W_v\), related to the volume ratio \(J\):

\begin{equation}
W_v(J) = \frac{\kappa}{2} \left( \frac{1}{2} (J^2 - 1) - \ln J \right)
\end{equation}

Differentiating \(W_v\), we obtain stress:

\begin{equation}
\tau_v = \frac{\partial W_v(J)}{\partial J} = \frac{\kappa}{2} (J - \frac{1}{J})
\end{equation}

Incorporating this stress into the Kelvin-Voigt model:

\begin{equation}
\sigma = \tau_v I + \eta \dot{\epsilon}
\end{equation}

Here, \(\tau_v\) multiplies the identity tensor \(I\), reflecting its isotropic effect.

As mentioned earlier, common fluid is hard to undergo significant volume change with the external force in household scenarios, in simulation, the volume-related stress is applied to keep the volume hold. Without this term, the volume will shrink in simulation.

\begin{figure*}[t!]
    \centering
    \includegraphics[width=0.9\linewidth]{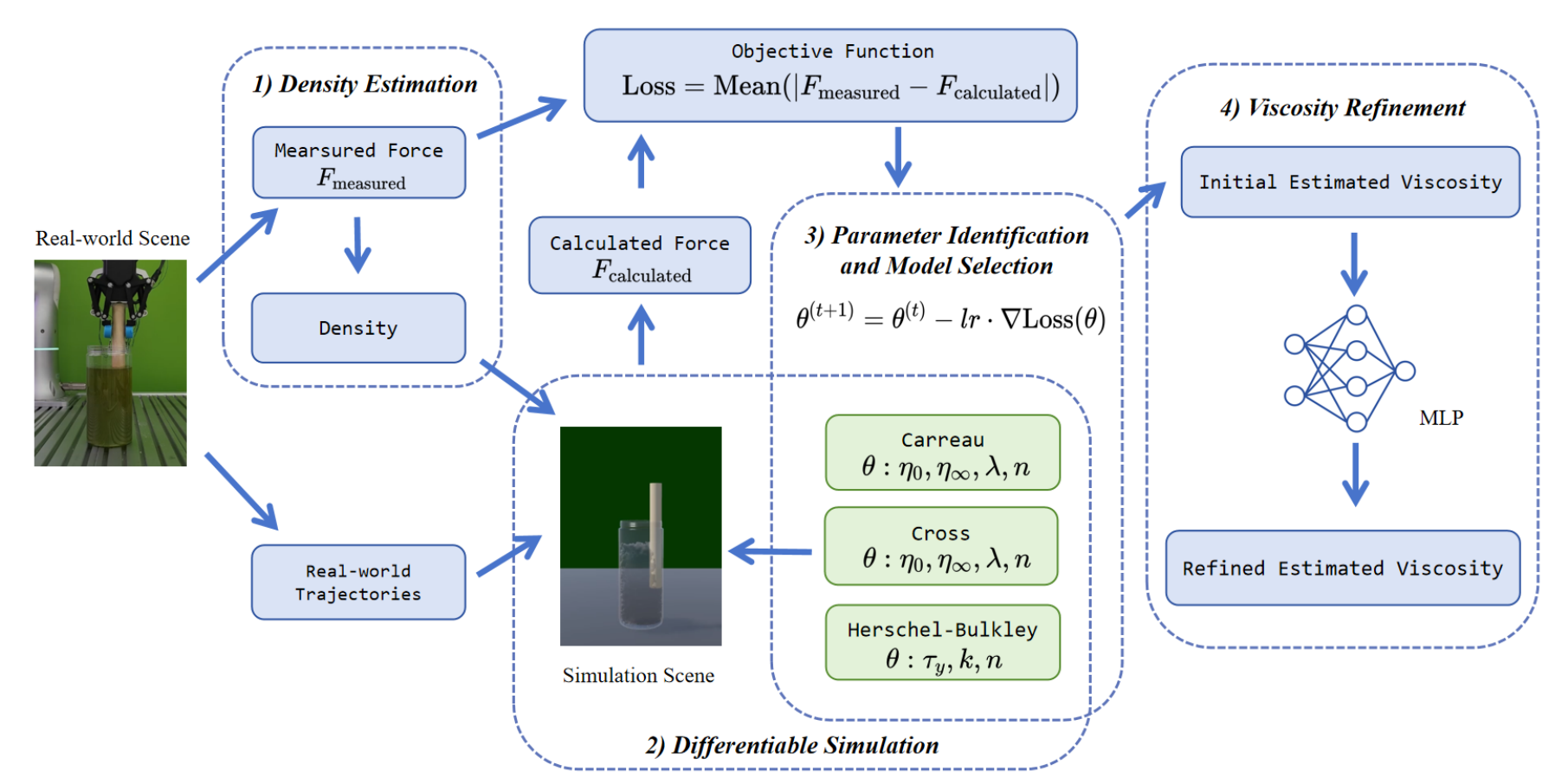}
    \caption{Pipeline of DiffStir. It takes the recorded real-world trajectories and the forces of a robot grasping a rod and stirring in a container. With the measured force, it can estimate the fluid density. The estimated density along with the trajectories can help to replicate the robotic stirring scene in a differentiable simulator. By comparing the force calculated from the simulator and the real-world measurements, it can adaptively find the most proper parameters and models. Finally, it adopts a neural network to refine the estimated parameters.}
    \label{fig:diffstir}
    \vspace{-0.3cm}
\end{figure*}

\textit{3) Extending to Non-Newtonian Fluids:}
To fully grasp how fluids behave, we combine models from continuum mechanics with non-Newtonian fluid models. Using this approach, we focus on three key non-Newtonian models: Carreau, Cross, and Herschel-Bulkley \cite{escudier2002fully}. These models help explain the changing relationship between viscosity \( \eta \) and shear rate \( \dot{\gamma} \) instead of treating viscosity as a fixed value. The function is named \textbf{apparent viscosity} and denoted as $\eta(\dot{\gamma})$. The detailed formulation of these models can be referred to in \cite{escudier2002fully} and our supplementary materials.

As shown in Fig. \ref{fig:diffstir}, the parameters of these models, denoted by \( \theta \) are: (a) $(\eta_0, \eta_\infty, \lambda, n)$ for the Carreau Model; (b) $(\eta_0, \eta_\infty, \lambda, n)$ for the Cross Model; (c) $(\tau_y, n, k)$ for the Herschel-Bulkley Model, where $\eta_0$ is the zero shear rate viscosity, $\eta_\infty$ is the infinite shear rate viscosity, $\lambda$ is a multiplier constant, $n$ is the power-law index, $\tau_y$ is the yield stress and $k$ is a consistency index. To note, the viscosity from the Herschel-Bulkley can be derived from the estimated yield. The formulation can be referred to in our supplementary materials.

We take the Carreau model as an example to demonstrate how the extended non-Newtonian formulation shapes. The apparent viscosity is given by:
\begin{equation}
\eta(\dot{\gamma})  = \eta_\infty + (\eta_0 - \eta_\infty) (1 + (\lambda \dot{\gamma})^2)^{(n-1)/2}
\end{equation}
And the fluid's stress, \( \sigma \), is a function of the parameters \( \theta = (\eta_0, \eta_\infty, \lambda, n) \):
\begin{equation}
\sigma(\theta) = \tau_v I + \left( \eta_\infty + (\eta_0 - \eta_\infty) (1 + (\lambda \dot{\gamma})^2)^{(n-1)/2} \right) \dot{\epsilon}
\end{equation}

\section{Method}
In this section, we describe the proposed DiffStir system.
Density and viscosity are estimated individually and sequentially. The overall pipeline is illustrated in Fig. \ref{fig:diffstir}.

\textit{1) Density Estimation:}
By integrating a force sensor at a robotic arm's end-effector, our method estimates density via force measurement on a submerged stirring rod. The density \( \rho \) is calculated with Archimedes' principle as:
\begin{equation}
    \rho = \frac{G-R}{V g} ,
\end{equation} 
where \( G \) is the gravitational force of the stirring rod, \( R \) is the equilibrium force measured by the force sensor, \( V \) is the volume of the immersed part of the rod, and the \( g \) is the gravitational acceleration. $G$, $R$, $V$ can all be measured from the real world, the experimental setup of measurement is described in Sec. \ref{sec:exp_setup_density}.

\textit{2) Differentiable Simulation:} To further estimate the viscosity, we build a bridge between the real and simulated observation via a differentiable simulator. We first replicate the setup of the robotic stirring in the real world to the simulator. The density estimated from the last step is used to set up the fluid in the simulated world. Then, we conduct the stirring operation in the real world, and transfer the recorded trajectories (e.g. rod positions, velocities) and the force applied to the rod to the simulated environment. The physics engine MLS-MPM \cite{hu2018moving} in the simulator can produce the interaction force between the fluid and the rod, $F_{calculated}$. 
The fluid is modeled by multiple particles, and the particles are moved by the stirring operation. The particle dynamics is governed by a constitutive model. Thus, the same stirring operation can cause different particle movements due to different constitutive models. If the constitutive model contains the parameter of viscosity, the particle movement can exhibit the effect of drag and horizontal shear. In this sense, the reactive $F_{calculated}$ is parametric.
In our simulation, we represent the fluid as a collection of particles. Each particle's stress, \( \sigma \), is governed by a constitutive relation. The force exerted on a single particle due to this stress can be computed as \(\mathbf{F}_{\text{particle}}(\theta) = \int \sigma(\theta) \cdot \mathbf{n} \, dA\), where \( \mathbf{n} \) is the outward normal of the particle's surface and \( A \) is the area over which the stress acts. The cumulative force exerted on the stirring rod, resulting from all particles in contact with it, is given by \( F_{\text{calculated}}(\theta) = \sum_{i=1}^{N} \mathbf{F}_{\text{particle}_i}(\theta) \). Here, \( N \) means the number of particles interacting with the rod.

\textit{3) Parameter Identification and Model Selection:}
We attach a thin-film pressure sensor to the gripper finger and measure the force from the rod, $F_{measured}$. By assuming no contact between the rod and the liquid container, the force applied to the pressure sensor is the same as the force applied to the rod from the fluid. In this way, we can compare the measured force $F_{measured}$ in the real world, and the calculated force $F_{calculated}$ in the simulator, and construct a loss term to optimize the parameters $\theta$ associated with $F_{calculated}$:
\begin{equation}\label{equ:loss}
    \text{Loss}(\theta) = \text{Mean}\left( | F_{\text{measured}} - F_{\text{calculated}}(\theta) | \right).
\end{equation}
By leveraging the power of differentiable simulation, optimization of \( \theta \) can be simply achieved using the gradient descent algorithm, expressed as:
\begin{equation}
\theta^{(t+1)} = \theta^{(t)} - lr \cdot \nabla \text{Loss}(\theta)
\end{equation}
\( lr \) denotes the learning rate, \( t \) is the iteration, and \( \nabla \text{Loss}(\theta) \) is the gradient of the loss function concerning \( \theta \).

Different constitutive models prefer specific fluids; thus, our system uses the Carreau, Cross, and Herschel-Bulkley models to suit different fluids. We employ an adaptive method to select the most suitable model and parameters, basing our choice on the lowest error from Eq. \ref{equ:loss} after three estimations. Notably, while these models predominantly describe non-Newtonian fluids, they can capture Newtonian behaviors under specific conditions, e.g., the Cross model when the power-law index \(n\) nears zero.

\textit{4) Viscosity Refinement:}
Given the numerical differences and measurement noises between simulations and real-world setups, our system uses a multi-layer perceptron (MLP) network to correct these systematic errors. We generate 2500 estimated/reference viscosity pairs as the training samples. We leave the details of the training set construction in supplementary materials.

Our MLP rectifier has an architecture that consists of two input neurons (representing average stirring rate and preliminary viscosity), a hidden layer with 8 neurons, and a single output neuron for refined viscosity.
For training, we set a learning rate of 0.01, achieving a balance between speed and stability in convergence. The model underwent 50 epochs of training, allowing for sufficient learning and weight adjustments. 

\section{Experimental Setup}

\subsection{Real-World Setup}\label{sec:real_setup}

To ensure rigorous experimentation for fluid dynamics during stirring, we design a precise setup, emphasizing alignment between simulation and real-world conditions (Fig. \ref{fig:diffstir}). 

\subsubsection{Setup Specifications}
We utilize a cylindrical PET container (diameter: 8.5cm, height: 16.5cm) and fill it with 738ml of fluid. A cylindrical wooden stirring rod (length: 20cm, diameter: 2.5cm) is submerged 8cm into the fluid.

\subsubsection{Force Sensing}
We sense the reactive force with an RP-C18.3-ST resistive thin-film pressure sensor with a diameter of 18.3mm on the robotic gripper, DH AG95, detecting force variations as minute as 0.005N.

\subsubsection{Calibration and Data Processing}
Before experimentation, a stationary ``dry run'' establishes a baseline force profile, accounting for non-fluidic forces. This baseline, when subtracted from stirring force data, reveals fluidic resistance.

\subsubsection{Tasks \& Fluid Selection}
We undertake four tasks: density validation, density \& apparent viscosity estimation, non-Newtonian fluid behavior validation, and fluid mixing analysis. Fluids range from Newtonian to non-Newtonian and span various viscosities and densities:

\textbf{Density Validation}: It is to validate the density estimation functionality alone. We employ high fructose syrup (F60), diluted syrup solution, water, and two starch solutions, which are prepared with ample starch to achieve high density.
    
\textbf{Density \& Apparent Viscosity Estimation}:  It is to validate the whole DiffStir pipeline. We employ corn starch solutions at 24 Baumé (CS 24) and 25 Baumé (CS 25), castor oil (Oil), propylene glycol (PG), ketchup, and water. To note, these fluids have larger variations in viscosity but fewer variations in density in comparison with the fluids used in the density validation task.

\textbf{Fluid Dynamics Behavior Classification}: It is to validate the sensitivity of our method to detect subtle fluid dynamics behavior such as Newtonian, shearing-thinning and shearing-thickening. We employ concentrated jam (Jam), corn starch solutions 24 Baumé (CS 24) and 25 Baumé (CS 25), castor oil (Oil), and concentrated coffee solution(Coffee).
    
\textbf{Fluid Mixing Analysis}: It is to validate the sensitivity of our method to detect subtle viscosity change during the mixing of two fluids. We employ high fructose syrup (F60) and water.

\subsection{Simulation Setup}

In the MLS-MPM simulation, we set the time step to \(10^{-3}\) seconds and the MPM grid width to 0.01m to ensure a balance between computational efficiency and simulation precision.  We use 6500 particles to balance fluid detail and computational speed within 24GB memory of Nvidia GeForce RTX 4090 GPU. Each simulation period runs for 8,000 frames, or 8 seconds, with a full gradient computation.

\subsection{Task Setup}

\subsubsection{Density Estimation \& Validation}\label{sec:exp_setup_density}

We measure the gravitational force \( G \) of the stirring rod using the end-effector's force sensor in the robot arm during ``dry run'' while keeping it vertical; the immersion depth \( h \) is measured with a 0.1mm-precision ruler; and the volume \( V \) is found by using \( V = A h \), where \( A \) is the cross-sectional area. When equilibrium is reached, the net force \( R \) is read directly from the end-effector's force sensor.

\subsubsection{Apparent Viscosity Estimation}

As shown in Fig. \ref{fig:viscosity_fluids}, we validate our system's viscosity estimations, \(\eta\), at a zero shear rate, \(\eta(\dot{\gamma} = 0)\), because many zero-shear results of different fluids are reported so that we can compare \cite{EngineeringToolboxDynamicViscosities, EngineeringToolboxViscosity, EngineeringToolboxKinematicViscosities}. If the apparent viscosity is consistent with the previous reports, we can validate the accuracy and usability of the estimated underlying physics parameters like \(\eta_0\), \(\eta_\infty\), \(\lambda\), \(\tau_y\), \(k\), and \(n\).%

\subsubsection{Fluid Dynamics Behavior Classification}
We investigate how viscosity estimations vary with changes in stirring speed, which directly influences the shear rate in the fluid simulation.

\subsubsection{Fluid Mixing Analysis}

We perform a mixing experiment to evaluate the system’s performance during stirring. %

We begin by blending F60 syrup and water. The initial viscosity of the F60 syrup, measured with the DiffStir system, is \(0.00770 \text{Pa$\cdot$s}\). Robot then stirs the mixture until fully combined, measuring viscosity at specific time intervals.

\section{Results}

\subsection{Density Validation}

We assessed the density estimation functionality alone using a fluid set with a large variation in density, including F60 syrup, diluted syrup solution, water, and two proportional starch solutions. The DiffStir estimates (``$\rho_{est}$'') were benchmarked against reference standards (``$\rho_{ref}$'')(calculated by mass/volume), as detailed in Table~\ref{tab:all_density_data_revised}.

\begin{figure*}[!t]
\vspace{0.2cm}
\centering
\includegraphics[width=0.9\linewidth]{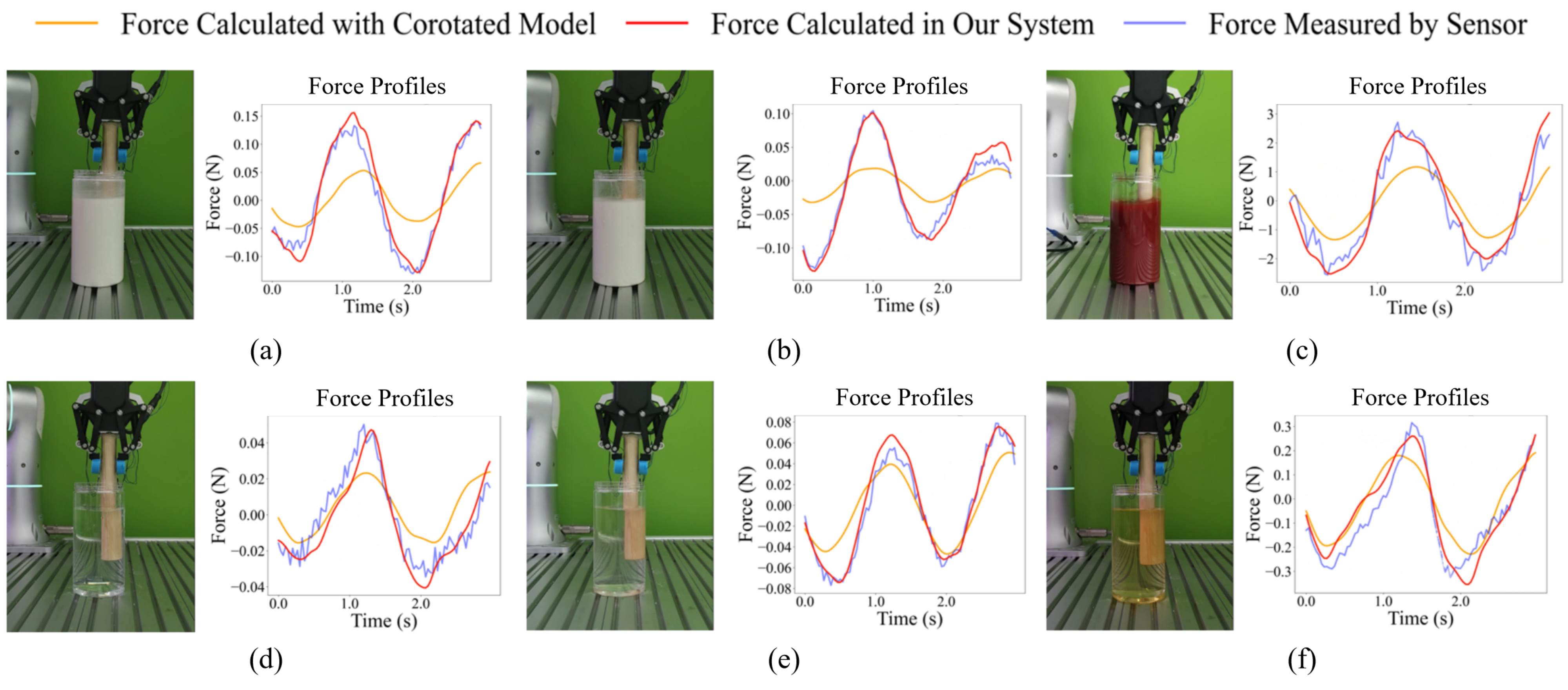}
\caption{Force Comparison During Stirring: From panels (a) to (f), the fluids represented are corn Starch solution (24 Baumé), corn starch solution (25 Baumé), ketchup, water, propylene glycol, and castor oil respectively. In each panel, the left side illustrates the real-world stirring scenario, and the right side showcases force comparisons in a fixed horizontal direction on the stirring rod. The red line depicts forces from simulations after fitting fluid parameters using our system, and the blue line, refined through adaptive Gaussian filtering, signifies experimental forces. The overlaps demonstrate DiffStir's precision.}
\label{fig:viscosity_fluids}
\vspace{-0.2cm}
\end{figure*}

\begin{table}[h]
\centering
\caption{Benchmarking DiffStir's density estimation functionality against reference standards (\( \text{kg/m}^3 \)).}
\label{tab:all_density_data_revised}
\begin{tabular}{l|ccc}
\toprule
\textbf{Fluid} & $\rho_{\text{est}}$ & $\rho_{\text{ref}}$ & \textbf{Bias (\%)} \\
\midrule
F60 Syrup & 1333.7 & 1290 & 3.39 \\
Diluted Syrup & 1065.8 & 1139.3 & -6.45 \\
Water & 915.1 & 1000 & -8.49 \\
Starch A & 1641.6 & 1563.7 & 4.98 \\
Starch B & 1814.8 & 1845.5 & -1.66 \\
\bottomrule
\end{tabular}
\end{table}

We calculate the bias by $\rho_{\text{est}}/\rho_{\text{ref}}-1$. As shown in Table \ref{tab:all_density_data_revised}, DiffStir system can provide accurate estimates (absolute bias $<5\%$) of most liquid types. The slightly higher bias observed for diluted syrup and Water can be attributed to their overall lower densities, making them more sensitive to measurement inaccuracies.

\subsection{Density \& Apparent Viscosity Estimation}
Since fluids with large variations in density may not necessarily have large variations in viscosity, we select a different set of fluids to process the viscosity estimations: corn starch solutions at 24 Baumé (CS 24) and 25 Baumé (CS 25), castor oil (Oil), propylene glycol (PG), ketchup, and water. To achieve this, we should also estimate the density of the fluid first for the simulation setup. In Table \ref{tab:density_before_viscosity_estimate}, we report all the parameters estimated for all the constitutive model concerns: the Carreau Model, Cross Model, and Herschel-Bulkley Model. 
Parameters for these models are reported with units: \(\tau_y\) in mPa, \(k\) in mPa\(\cdot\)s\(^n\), \(\eta_0\) and \(\eta_\infty\) in mPa\(\cdot\)s, \(n\) as dimensionless, and \(\lambda\) ensuring dimensional consistency in the respective formulas. 

\begin{table}[h]
\footnotesize
\centering
\caption{Estimated density (\( \text{kg/m}^3 \)) and parameters for the Carreau, Cross, and Herschel-Bulkley models for various fluids.}
\label{tab:density_before_viscosity_estimate}
\begin{tabular}{p{0.7cm}p{0.8cm}p{1.6cm}p{1.6cm}p{1.8cm}}
\toprule
\textbf{Fluid} & \textbf{$\rho_{\text{est}}$} & \textbf{Carreau \newline \(\eta_0, \eta_\infty,\)\newline \(n, \lambda\)} & \textbf{Cross \newline \(\eta_0, \eta_\infty,\)\newline\( n, \lambda\)} & \textbf{Herschel-Bulkley \newline \(\tau_y, k, n\)}\\
\midrule
CS 24 & 1146.82 & 224, 756, \newline1.1, 0.023 & 255, 812,\newline 0.35, 0.1 & 341, 214, 1.3 \\\hline
CS 25 & 1175.38 & 903, 1347,\newline 1.1, 0.002 & 957, 1342,\newline 0.31, 0.04 & 335, 858, 1.3 \\\hline
Oil & 866.88 & 1794, 341,\newline 1.0, 16 & 1667, 1977,\newline 0.07, 1231 & 23, 1868, 0.8 \\\hline
PG & 936.27 & 16, 66, \newline1.1, 0.026 & 24, 76,\newline 0.00, 93 & 11, 33, 0.8 \\\hline
Ketchup & 1156.1 & 3989, 2567,\newline 0.7, 5992 & 6135, 5568,\newline 0.70, 1.97 & 2663, 3442, 0.5\\\hline
Water & 915.1 & 3, 5,\newline 1.1, 0.001 & 4, 53,\newline 0.01, 9.68 & 17, 0.01, 1.1\\
\bottomrule
\end{tabular}
\end{table}

According to our adaptive selection strategy, we will automatically select the most proper model to characterize the fluids, as shown in Table \ref{tab:force_deviation}. It delineates average force deviations (N) for different fluids over 8s force curves. The model manifesting the smallest deviation, marked in bold, is selected as the most proper for providing accurate viscosity estimations. The system generally shows small force deviations, with the most precise measurement reaching as low as 0.006 N for PG.

\begin{table}[h]
\centering
\caption{Average Force Deviations (N) for Various Fluids using Different Models.}
\label{tab:force_deviation}
\begin{tabular}{l|ccc}
\toprule
\textbf{Fluid} & \textbf{Carreau} & \textbf{Cross} & \textbf{Herschel-Bulkley} \\
\midrule
Ketchup & 0.475 & 0.513 & \textbf{0.397} \\
CS 24 & \textbf{0.010} & 0.016 & 0.255 \\
CS 25 & \textbf{0.017} & 0.023 & 0.347 \\
Oil & \textbf{0.048} & 0.063 & 0.175 \\
PG & 0.008 & \textbf{0.006} & 0.151 \\
Water & 0.011 & \textbf{0.009} & 0.137 \\
\bottomrule
\end{tabular}
\end{table}

We also justify our choice of the constitutive models against the baseline model, the corotated model, which is adopted in FluidLab \cite{xian2023fluidlab}, on reactive force comparison and viscosity estimation. Table~\ref{tab:dynamic_viscosity_data} presents a comparison between viscosity estimated with our model, those with the Corotated model, and the reference values from established scientific principles and empirical findings \cite{EngineeringToolboxDynamicViscosities, EngineeringToolboxViscosity, EngineeringToolboxKinematicViscosities}.
Fig. \ref{fig:viscosity_fluids} illustrates the congruence between simulation-calculated forces and real-world measurements, underscoring DiffStir's capability to accurately reflect real-world dynamics based on its adapted fluid constitutive parameters.

\begin{table}[ht!]
\centering
\caption{Comparison of Apparent Viscosity Values from Our System: Using our constitutive model and the baseline corotated model, against reference values.}
\label{tab:dynamic_viscosity_data}
\begin{tabular}{l|ccc}
\toprule
\textbf{Fluid} & \textbf{Our / Corotated} & \textbf{Reference} & \textbf{Bias (\%)} \\
\midrule
CS24 & 133 / 87 & 155.55 & -14.50 / -44.07 \\
CS25 & 338 / 215 & 366.25 & -7.71 / -41.30 \\
Oil & 615 / 743 & 650 & -5.38 / 14.31 \\
PG & 18.7 / 11.7  & 16.2 & 15.43 / -27.78 \\
Ketchup & 2755.3 / 2145  & 3000 & -8.16/ -28.50 \\
Water & 0.73 / 0.58 & 0.89 & -17.98 / -34.83 \\
\bottomrule
\end{tabular}
\end{table}

Finally, we validate the effectiveness of the viscosity refinement.
Table \ref{tab:Comparative_viscosity_estimations} contrasts the initial and refined estimations with the reference values.
We evaluate the effectiveness of our refinement by comparing the error metrics of the initial and refined models. We use the mean absolute error (MAE) and the root mean square error (RMSE) to measure the accuracy of the apparent viscosity estimation. The results show that incorporating the MLP rectifier significantly reduces the error metrics and improves the estimation accuracy. The initial model has an MAE of 348.80mPa$\cdot$s and an RMSE of 534.11mPa$\cdot$s, while the refined model has an MAE of 55.53mPa$\cdot$s and an RMSE of 101.99mPa$\cdot$s. 

\begin{table}[ht!]
\centering
\caption{Comparative viscosity estimations.}
\label{tab:Comparative_viscosity_estimations}
\begin{tabular}{l|ccc}
\toprule
\textbf{Fluid} & \textbf{Initial} & \textbf{Refined} & \textbf{Reference} \\
\midrule
CS24 & 223 & 133 & 155.55 \\
CS25 & 902 & 338 & 366.25 \\
Oil & 1793 & 615 & 650 \\
PG & 23 & 18.7 & 16.2 \\
Ketchup & 2663 & 2755.3 & 3000 \\
Water & 3.7 & 0.73 & 0.89 \\
\bottomrule
\end{tabular}
\end{table}

\subsection{Fluid Dynamics Behavior Classification}

\begin{figure}[htbp]
    \centering
    \includegraphics[width=1\linewidth]{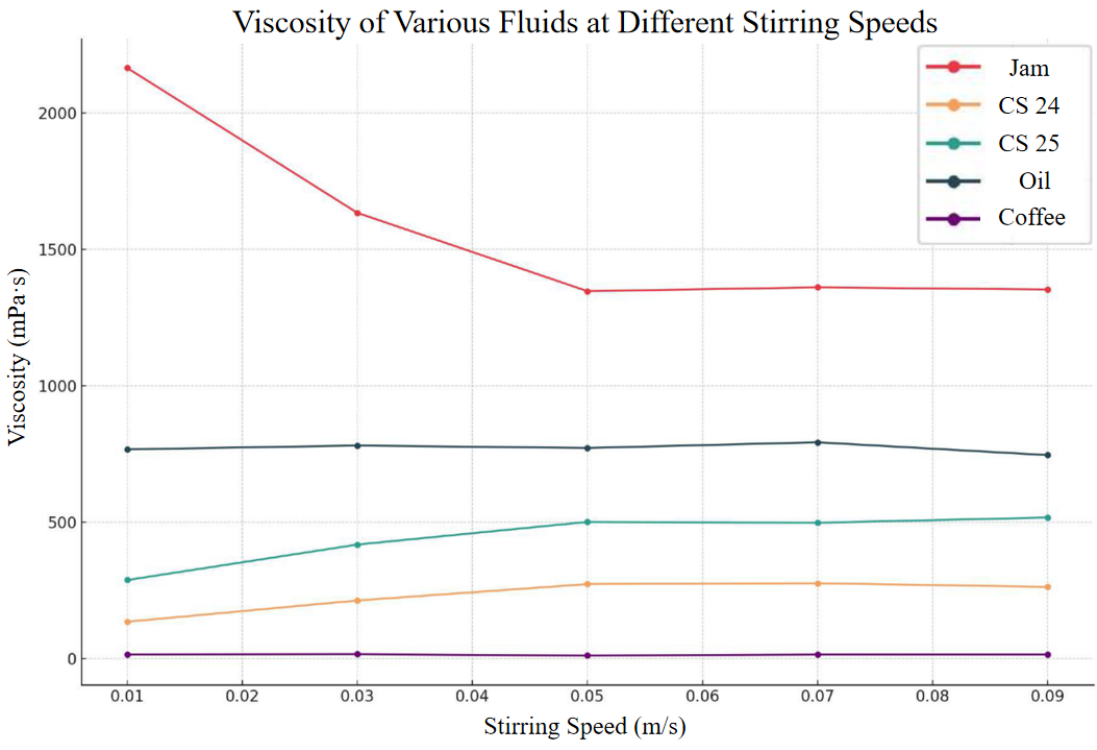}
    \caption{Viscosity trends across stirring speeds for multiple fluids. Markers depict data points, while lines show the viscosity variation (in mPa$\cdot$s) with stirring speed (in m/s).}
    \label{fig:viscosity_data}
\end{figure}

Viscosity variations for fluids, such as concentrated jam (Jam), corn starch solutions 24 Baumé (CS 24) and 25 Baumé (CS 25), castor oil (Oil), and concentrated coffee solution (Coffee), were assessed across a spectrum of stirring rates. The results are presented in Fig. \ref{fig:viscosity_data}.

As the stirring speed increases, correspondingly the shear stress increases, the viscosity will have three different evolutional directions: decrease stands for shear thinning (concentrated jam), increase stands for shear thickening (CS 24 and CS 25), and maintaining stands for Newtonian behavior (Oil and Coffee).

\begin{figure}[t!]
    \centering
    \includegraphics[width=1\linewidth]{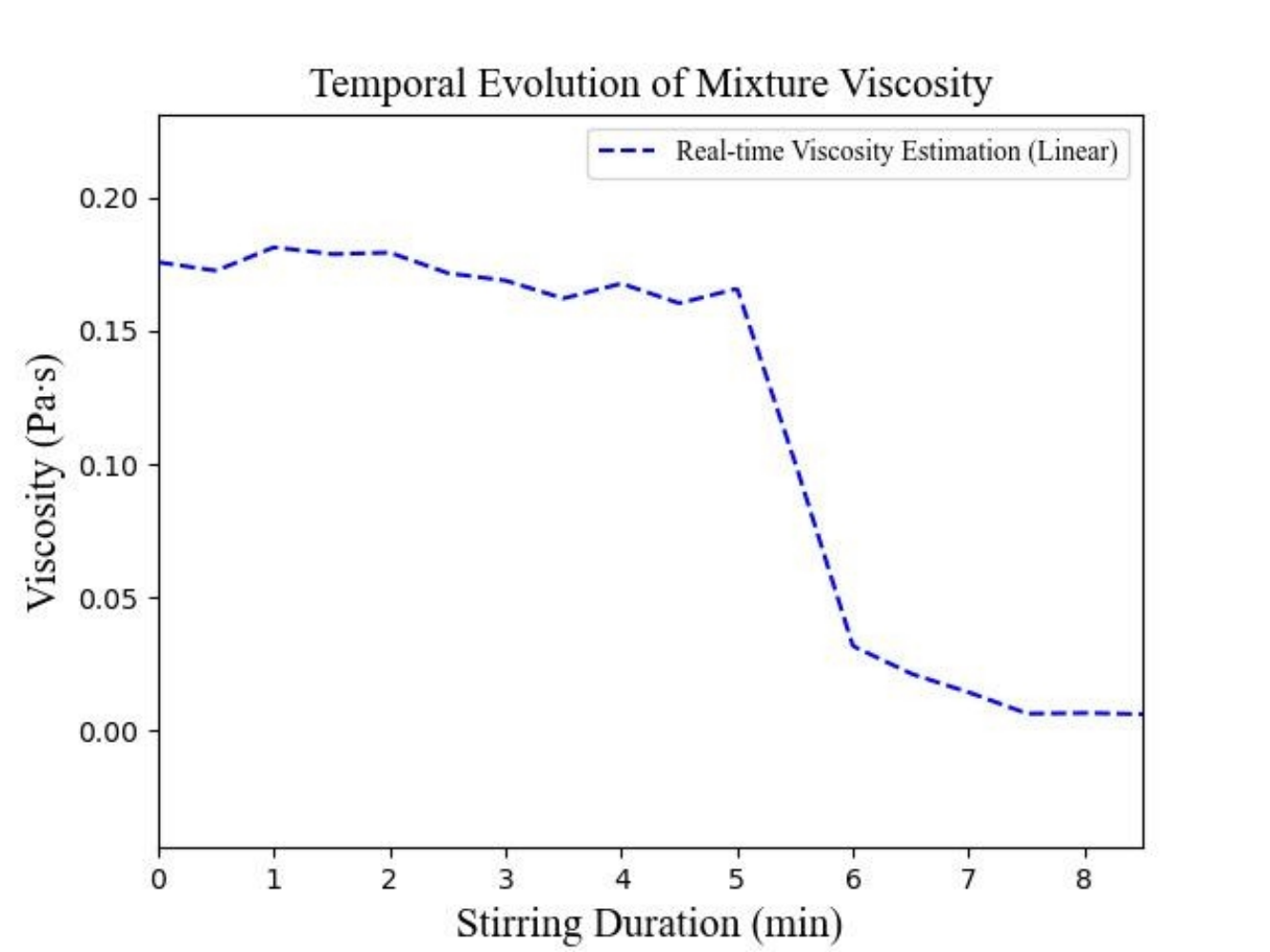}
    \caption{Temporal evolution of viscosity with stirring duration.}
    \label{fig:viscosity_change}
\end{figure}

\subsection{Fluid Mixing Analysis}

Fig.~\ref{fig:viscosity_change} shows a sharp decrease in viscosity as stirring duration increases, which is expected when combining high-viscosity F60 syrup with lower-viscosity water. Initially, the blend's viscosity is mainly dictated by the F60 syrup due to its higher density and contact with the rod. As stirring continues, the two liquids disperse to each other and eventually reach equilibrium. In this process, we can observe the viscosity drop and reach a plateau in the end.

The mixed viscosity is calculated by the Lederer-Roegiers equation \cite{zhmud2014viscosity}, which is a practical tool in rheology to predict the viscosity \(\eta_{\text{blend}}\) of a mixture by considering the individual viscosities \(\eta_{1}\) and \(\eta_{2}\) of two fluids, and the volume fraction \(\phi\) of the first fluid, formulated as \(\eta_{\text{blend}} = \eta_{1}^{(\alpha \phi)} \cdot \eta_{2}^{[(1-\alpha) (1-\phi)]}\). The empirical parameter \(\alpha\), typically varying between 0 and 1, fine-tunes the equation, capturing the influences of each fluid on the mixture's overall behavior. 

Using the Lederer-Roegiers equation, predicted viscosities for the blend at empirical parameter \(\alpha\) values of 0.5 and 1.0 are \(0.01428\text{Pa·s}\) and \(0.00527\text{Pa·s}\) respectively, which establish the theoretical viscosity range spans of this mixture. Our experimentally estimated blend viscosity (\(0.00770\text{Pa·s}\)), aligns well within this spectrum, highlighting the accuracy of our approach.

\section{Conclusion}
In this work, we introduced DiffStir, a novel differentiable fitting framework, adept at identifying the key physics parameters through a commonly practiced action, stirring. By using a robotic arm and synchronizing real-world actions with a diffMPM-based simulator, the density and viscosity of various fluids are estimated with considerable accuracy. We've showcased that not only can the method adapt to different fluid dynamics behaviors but also has the potential to bridge the gap between physical simulations and real-world observations. Through comprehensive evaluations, DiffStir has exhibited its efficiency and precision in parameter estimation, marking a notable advancement in the domain of fluid dynamics understanding in everyday scenarios.

\bibliographystyle{IEEEtran}
\bibliography{root}

\end{document}